\def\year{2019}\relax
\documentclass[letterpaper]{article} 
\usepackage{aaai19}  
\usepackage{times}  
\usepackage{helvet}  
\usepackage{courier}  
\usepackage{url}  
\usepackage{graphicx}  
\usepackage{color}
\usepackage{multirow}
\usepackage{amsmath}
\usepackage{amssymb}
\usepackage{verbatim}
\usepackage[ruled]{algorithm2e}

\newcommand{\eg}{\emph{e.g. }}
\newcommand{\ie}{\emph{i.e. }}
\newcommand{\etal}{\emph{et al. }}

\begin{document}
%
\title{STA: Spatial-Temporal Attention for Large-Scale Video-based Person Re-Identification}

\author{Yang Fu$^{1,2}$\thanks{Yang Fu did this work during the intern at Nokia Bell Labs. Xiaoyang Wang and Yunchao Wei are joint corresponding authors.}, Xiaoyang Wang$^{2}$, Yunchao Wei$^{1}$, Thomas Huang$^{1}$ \\
{\small $^1$IFP, Beckman, UIUC, IL $^2$Nokia Bell Labs, Murray Hill, NJ}\\
{\tt\small \{yangfu2, yunchao, t-huang1\}@illinois.edu} \\
{\tt \small xiaoyang.wang@nokia-bell-labs.com}
}
\renewcommand\footnotemark{}

\maketitle
\thispagestyle{empty} 
\begin{abstract}
In this work, we propose a novel Spatial-Temporal Attention (STA) approach to tackle the large-scale person re-identification task in videos. Different from the most existing methods, which simply compute representations of video clips using frame-level aggregation (\eg average pooling), the proposed STA adopts a more effective way for producing robust clip-level feature representation. Concretely, our STA  fully exploits those discriminative parts of one target person in both spatial and temporal dimensions, which results in a 2-D attention score matrix via inter-frame regularization to measure the importances of spatial parts across different frames. Thus, a more robust clip-level feature representation can be generated according to a weighted sum operation guided by the mined 2-D attention score matrix. In this way, the challenging cases for video-based person re-identification such as pose variation and partial occlusion can be well tackled by the STA. We conduct extensive experiments on two large-scale benchmarks, \ie MARS and DukeMTMC-VideoReID. In particular, the mAP reaches {\bf 87.7\%} on MARS, which significantly outperforms the state-of-the-arts with a large margin of more than {\bf 11.6\%}.
\end{abstract}
\section{Introduction}

Person re-identification (Re-ID) aims at matching images of a person in one camera with the images of this person from another different camera. In recent years, person Re-ID under the image sequence (video) setting has drawn significant attention in literature since it is critical for applications like intelligent video surveillance and multimedia.~\cite{mclaughlin2016recurrent,wang2016person,zhou2017see,liu2017quality,xiao2019ian}.
\begin{figure*}[t]
	\centering
	\footnotesize
	\includegraphics[width=0.75\textwidth]{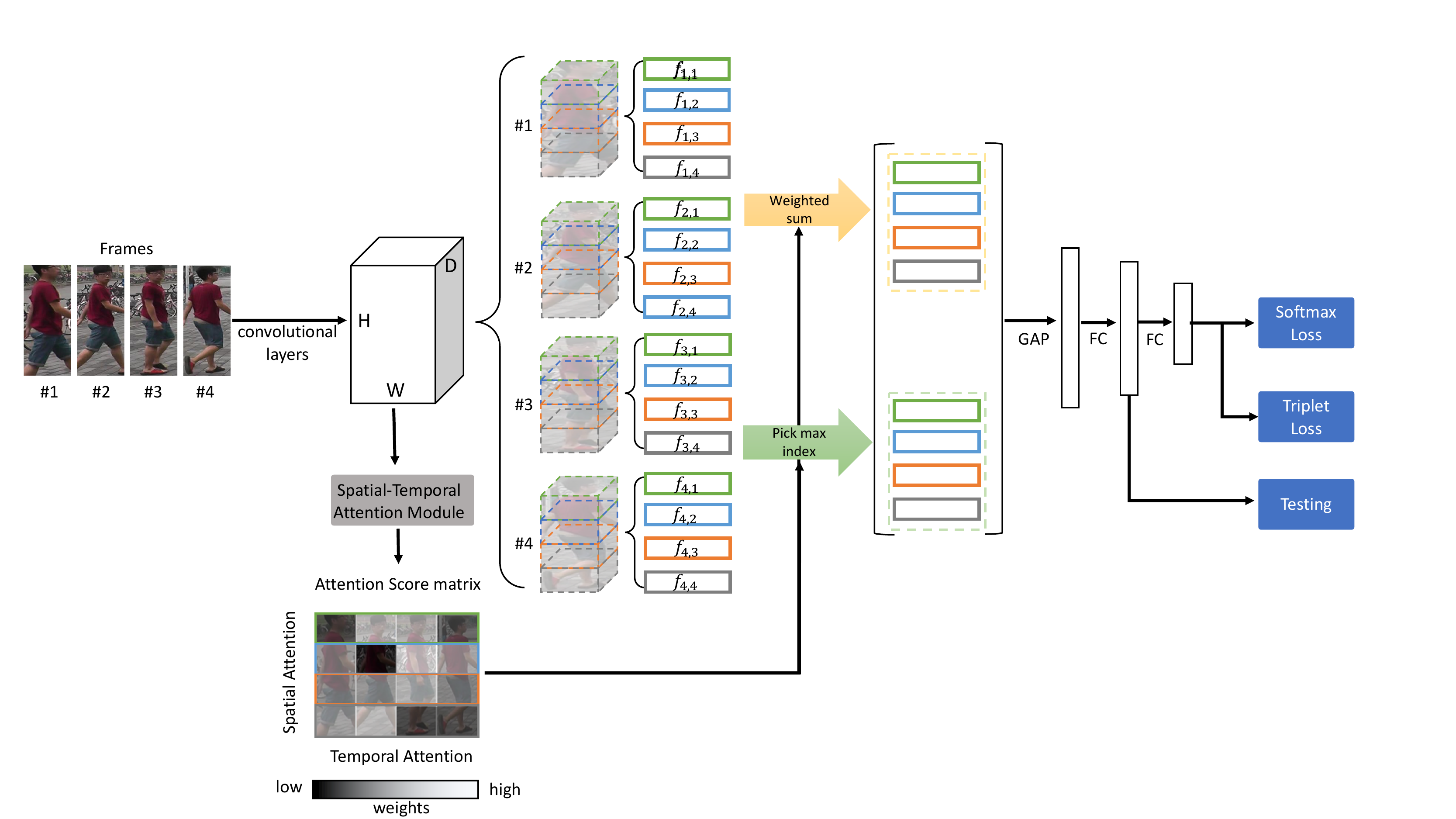}
	\caption{{\bf Architecture of STA framework.} The input video tracklet is first reduced to $N$ frames by random sampling. (1) Each selected frame is fed into the backbone network to be transformed into feature maps. (2) Then, the feature maps are sent to our proposed spatial-temporal attention model to assign an attention score to each spatial region of different frames and then generate a 2-D attention score matrix. An Inter-Frame Regularization is employed to restrict the difference among different frames (detailed in Fig.~\ref{fig:sta}). (3) Using the attention score, we extract the spatial region feature map with the highest attention score among all frames, and operate the weighted sum of the spatial region feature maps based on the assigned attention score. (4) Then, we adopt feature fusion strategy to concatenate the spatial feature maps from different spatial regions to generate two sets of feature maps of the whole person body as the global representation and discriminative representation. (5) Finally, a global pooling layer and a fully connected layer are used to transform the feature maps to a vector for person re-identification. During training, we combine both the triplet loss and the softmax loss. During testing, we choose the feature vector after the first fully connected layer as the representation for the input video tracklet.}
	\vspace{-5mm}
	\label{fig:architecture}
\end{figure*}

Most of the existing video-based person Re-ID work~\cite{liu2017quality,song2017region,li2018diversity} focuses on very small datasets, \eg PRID-2010~\cite{hirzer2011person} and iLIDS-VID~\cite{wang2014person} that only contains about 300 person identities with 600 tracklets in total. Although the existing approaches achieve good performances on PRID-2010 and iLIDS-VID, their accuracies in the state-of-the-art large-scale datasets such as the MARS dataset~\cite{zheng2016mars} and DukeMTMC-VideoReID dataset~\cite{wu2018cvpr_oneshot,ristani2016MTMC} are far from satisfying. Their performances are largely limited by the tremendous variations in camera viewpoints, human poses, illumination, occlusions, and background clutter in the large-scale video-based Re-ID dataset.

For the video-based Re-ID task, the key is to learn a mapping function that converts videos into a low-dimensional feature space in which each video can be represented by a single vector. Most existing methods represent a frame of the video as a feature vector, and then employ average or maximum pooling across the frames to obtain the representation of the input video~\cite{hirzer2011person,you2016top}. However, this approach usually fails when occlusions appear frequently in the video. In addition, basic operations like maximum or average pooling across video frames cannot handle the spatial misalignment caused by the variation of human poses among frames. In order to distill the relevant information from a video and weaken the influence of noisy samples (\eg occlusion), recent studies~\cite{zhou2017see,xu2017jointly} introduce the attention mechanism and achieve improved results. However, these existing attention-based methods only assign an attention weight to each frame, thus lack the capabilities of discovering either the discriminative frames in a video sequence or the discriminative body parts in each frame. Also, the attention mechanisms in most existing attention-based approaches are parametric, {\eg fully connected layers}, which require the input video sequence to be in fixed length.

In order to address the issues mentioned above, we propose an effective yet easy-to-implement \textbf{Spatial-Temporal Attention (STA)} framework to address the large-scale video-based person re-identification problem, as shown in Fig.~\ref{fig:architecture}. Instead of simply encoding a sequence of images by pooling or assigning weights to each frame by the parametric model, our STA framework jointly incorporates multiple novel components including frame selection, discriminative parts mining, and feature aggregation without using any additional parameters. In all, our major contributions in this paper can be summarized as follows:
\begin{itemize}
\item We propose a simple yet effective STA model that assigns attention score for each spatial region to achieve discriminative parts mining and frame selection without using any additional parameters.
\item We introduce the inter-frame regularization term to restrict the dissimilarity among different frames and ensure that each frame shares the same identity.
\item We design a novel feature fusion strategy that combines both global information and discriminative information from the video sequence for better feature aggregation.
\item We conduct extensive experiments and ablation study to demonstrate the effectiveness of each component. The final results achieve the state-of-the-art on two mainstream large-scale datasets: MARS and DukeMTMC-VideoReID.
\end{itemize}

\section{Related Work}
In this section, we review the related work including the image-based person re-identification, the video-based person re-identification, and the attention mechanism used in person re-identification.

\textbf{Image-based person re-identification} is extensively explored in the literature, and the existing studies can generally be divided into two categories: discriminative learning~\cite{xiao2016learning,zheng2017discriminatively,sun2017beyond,fu2018horizontal} and metric learning~\cite{ahmed2015improved,hermans2017defense,ding2015deep}. In \cite{hermans2017defense}, Hermans~\etal propose a variant of triplet loss to perform end-to-end deep metric learning, and their model can outperform many other published methods by a large margin. In \cite{zheng2017discriminatively}, Zheng~\etal employ siamese network and combine both verification loss and classification losses to learn a discriminative embedding and a similarity measurement at the same time.

\textbf{Video-based person re-identification} is an extension of the image-based person re-identification, and is widely studied recently~\cite{mclaughlin2016recurrent,liu2017video,wang2016person,zheng2016mars}. For example, McLaughlin~\etal \cite{mclaughlin2016recurrent} adopt Recurrnet Neural Network(RNN) to pass the message of each frame extracted from Convolution Network Network(CNN). Liu~\etal \cite{liu2017video} focus on learning a long-range motion context features from adjacent frames for a robuster identification.

\textbf{Attention models in person re-identification.} Since Xu~\etal \cite{xu2015show} propose the attention mechanism, it has been applied to lots of person re-identification work~\cite{zhou2017see,xu2017jointly,liu2017quality,li2018diversity}. In~\cite{liu2017quality}, Liu~\etal propose a method to estimate quality score of each frame automatically and weaken the influence of noisy samples. Xu~\etal \cite{xu2017jointly} introduce the joint Spatial and Temporal Attention Pooling Network that can extract the discriminative frames from probe and gallery videos, and obtain temporal attention weights for one sequence guided by the features of the other sequences. Li~\etal \cite{li2018diversity} employ multiple spatial attention models with the temporal attention model to learn latent representations of different body parts of each person. Comparing to these existing attention models, our proposed STA model has two major differences that further boost our person Re-ID performance: First, the simple yet effective STA model has no additional parameters, which means that the length of the input sequence does not have to be fixed. Second, the STA model can learn a attention score for each region in different frames, which can achieve the discriminative region mining and frame selection jointly.

\section{Proposed Method}
Given a tracklet of person sequence, we propose the STA framework (Fig.~\ref{fig:architecture}) to better handle the video-based person re-identification problem through spatial-temporal attention model with inter-frame regularization. We first randomly select a constant number of frames from the input video query, and feed them into a backbone network to extract features from each frame. Then, we send the obtained feature maps into the proposed STA model to generate a 2-D attention score matrix which assigns an attention weight for each spatial region of each frame. In order to restrict the difference among the frames in the single video tracklet, we propose the inter-frame regularization to estimate the inter-frame similarity. Next, we use both the spatial region with the maximum corresponding weights in each frame, and the weighted sum among all attention weights to obtain two sets of feature maps of the whole person body. Finally, we concatenate them together as the global representation and discriminative representation, and employ a global average pooling followed by a fully connected layer to represent the video query. For the objective function, we combine the softmax loss and batch-hard triplet loss~\cite{hermans2017defense}.

\subsection{Spatial-Temporal Attention (STA) Framework}
{\bf Backbone Network.} Various network architectures, like VGG~\cite{simonyan2014very}, Resnet~\cite{he2016deep}, and Google Inception~\cite{szegedy2016rethinking}, can be employed as the backbone network to extract feature maps for each frame. We choose ResNet50~\cite{he2016deep} as the backbone network, which was adopted by most previous works. In particular, ResNet50 has one convolutional block named $conv1$, and followed by four residual block named $conv2,3,4,5$ respectively. We further make two modifications on the original ResNet50: 1) the stride of the first residual block $conv5$ is set to $1$; 2) the average pooling layer and fully connected layer are removed. The input video is first reduced to $N$ frames by random sampling, and each selected frame is fed into the backbone network. As a result, each video $V=\{I_1,...,I_n,...,I_N\}$ is represented by a set of $16  \times 8$ feature maps $\{f_{n}\}_{\{n=1:N\}}$, and each feature map has $D=2048$ channels.

{\bf Spatial-Temporal Attention Model}.
We propose the spatial-temporal attention model to automatically learn from each image frame the discriminative regions that are useful for re-identification. Previous video-based person re-identification methods~\cite{liu2017quality,zhou2017see} consider each frame as a whole image and assign one weight for each frame. However, different regions of a person body should have different influences on the re-identification task. Thus, our approach aims to discover the discriminative representation of these region for each frame. Li~\etal \cite{li2018diversity} also employ a spatial-temporal attention model, in which they use different convolutional layers to extract salient region of person body and adopt traditional temporal attention model for frame selection. There are three major drawbacks of this method. First, it involves more computation because of more convolutional layers, and its input sequence length has to be fixed due to the temporal attention model. Second, multiple spatial attention models used in their approach are independent from each other, without utilizing the spatial relationships that exist between human body parts. As a result, the extracted spatial attentions could be scattered and do not reflect the complete human body in the foreground. Third, the spatial attention information and temporal attention information are obtained by two different models which would cause error accumulation. Different from the existing methods, our spatial-temporal attention assigns attention weights, which contain both of spatial attention information and temporal attention information, to each spatial region in different frames automatically without any additional parameters. Experiments in Table~\ref{exp:mars} demonstrate the advantages of our method compared to \cite{li2018diversity}. To the best of our knowledge, our model is the first video-based person Re-ID model that can discover the discriminative parts but reserve the spatial relationship, and achieve frame selection at the same time.

\begin{figure}[t]
	\centering
	\footnotesize
	\includegraphics[width=0.5\textwidth]{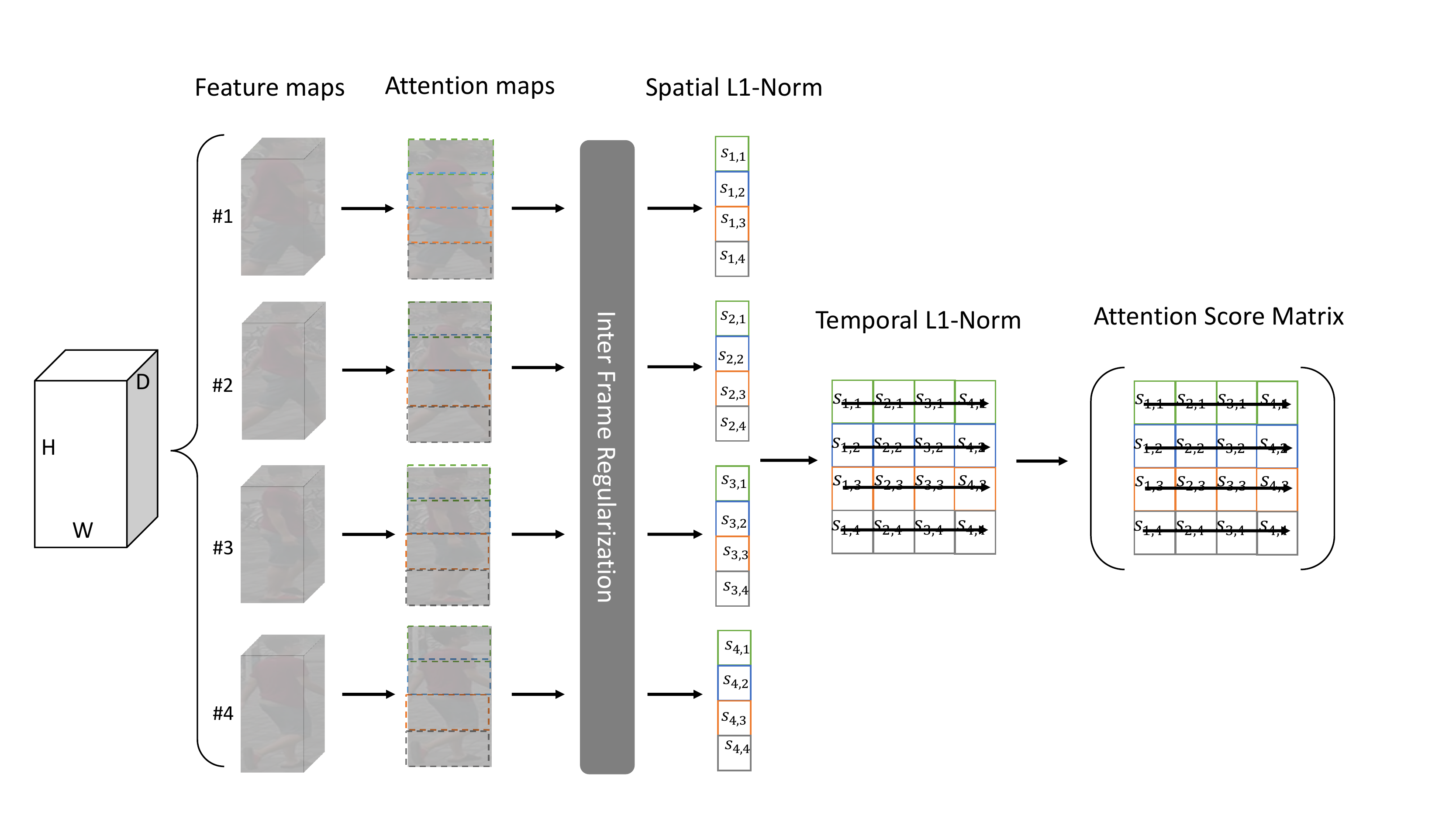}
	\caption{ {\bf Details of the spatial-temporal attention model with inter-frame regularization.} Given a set of feature maps from the input video, we generate corresponding attention maps for each frame. The inter-frame regularization is used to restrict the difference among frames in the same video tracklet. Then, attention maps are horizontally split into four equal spatial region, and the spatial region from the same spatial region but different frames are used to calculate the 2-D attention score matrix}
    \vspace{-5mm}
	\label{fig:sta}
\end{figure}

The illustration of the spatial-temporal attention model is shown in Fig.~\ref{fig:sta}. Given the feature maps of an input video $\{f_{n}\}_{\{n=1:N\}}$, we first generate the corresponding attention map $g_{n}$ by operating the $\ell_2$ normalization on the square sum through the depth channel. Specifically,
\begin{equation}\label{eq:attentionmap}
g_{n}(h, w) = \frac{||\sum_{d=1}^{d=D}f_{n}(h, w, d)^2||_2}{\sum_{h,w}^{H, W}||\sum_{d=1}^{d=D}f_{n}(h, w, d)^2||_2}
\end{equation}
where $H,W$ are the height and width of feature maps. Thus, each frame has one corresponding attention map. Both the feature maps and the attention maps of the $N$ frames are then divided into $K$ blocks horizontally:
\begin{equation}
\left\{
\begin{aligned}
    g_{n}=& [g_{n,1},..., g_{n,k},..., g_{n,K}] \\
    f_{n}=& [f_{n,1},..., f_{n,k},..., f_{n,K}]
\end{aligned}
\right.
\end{equation}
Here, $g_{n, k}$ represents the spatial attention map on $k$th regions of $n$th frame. After that, we employ the $\ell_1$ normalization on all values in each block to obtain one spatial attention score for that region.
\begin{equation}
    s_{n,k}=\sum_{i,j}||g_{n,k}(i,j)||_{1}
\end{equation}
Since the feature maps are after ReLU activation and all the values are greater than or equal to zero, the higher response of an attention map means the better representation of the person for re-identification task. The same procedure is operated on all selected frames of the input video to obtain the $N\times K$ matrix $S$ of spatial attention scores.

Instead of using multiple convolutional layers to formulate the temporal attention model as in \cite{li2018diversity,zhou2017see}, we directly compare the attention scores that are from different frames but on the same spatial region, and compute each attention over the $\ell_1$ normalization among them to obtain the normalized spatial-temporal attention scores. Specifically,
\begin{equation}
    S(n,k)=\frac{s_{n,k}}{\sum_{n}||s_{n,k}||_1}
\end{equation}
As a result, each spatial region from different frames is assigned with a specific attention score based on the spatial-temporal attention information.

\subsection{Inter-Frame Regularization}
For the video-based person Re-ID, the images from the same video tracklet of a person should represent the appearance of the same person. Such information is further exploited by our approach as the inter-frame regularization to restrict the difference of the learned attention maps among frames. This inter-frame regularization helps to avoid the cases in which the learned attention scores of each spatial region focus on one specific frame and largely ignore the other frames.

Specifically, since each frame has a corresponding feature map $f_{n}$, which is used to classify the person identification during training. One possible way is to add a classification loss to all frames to make sure they share the same identification. However, there maybe some noisy samples which are hard to classify and hence make the training processing unstable. An alternative solution is to use Kullback-Leibler (KL) divergence to evaluate the similarity of each frame, but lots of close-to-zero elements exist in attention maps. These elements will drop dramatically when employing the $log$ operation in the KL divergence, and make the training processing unstable as well~\cite{lin2017structured}. Thus, in order to encourage the spatial-temporal attention model to preserve the similarity and meanwhile avoid focusing on one frame, we design the inter-frame regularization which measures the difference among input image frames. For convenience, we define $G$ as the collection of attention maps generated from the input image frames,
\begin{equation}
G = [g_1, ..., g_N]
\end{equation}
Assume $g_i,g_j$ are attention maps calculated by Eqn. (\ref{eq:attentionmap}) for two frames $i$ and $j$. We employ the square Frobenius Norm~\cite{meyer2000matrix} of the difference between $g_i$ and $g_j$. Specifically,
\begin{equation}
\begin{aligned}
Reg & = ||g_i - g_j||_F \\
& = \sqrt{\sum_{h=1}^{H}\sum_{w=1}^{W}|g_i(h, w)-g_j(h, w)|^2}
\end{aligned}
\end{equation}
Note that we randomly choose two frames $i$ and $j$ from the $N$ frames of each video for this regularization term. In order to restrict the difference between two frames, we minimize this regularization term $Reg$ by adding it to the original objective function $L_{total}$ defined in Eqn. (\ref{eq:loss}) after multiplied by a coefficient $\lambda$.
\begin{equation}
    \min(L_{total} + \lambda Reg)
\end{equation}

\subsection{Feature Fusion Strategy}
After STA model with inter-frame regularization, we obtain an $N \times K$ matrix $S$ that assigns an attention score $s_{n,k}$ for the feature map $f_{n,k}$ of each spatial region and each frame. Inspired by~\cite{fu2018horizontal}, we propose a strategy for feature fusion by combining the global and discriminative information of each tracklet as described in Alg.~\ref{algo:a1}.

Given the attention score matrix and a set of feature maps, we first divide feature maps into several spatial regions just like what we operate on attention map, and pick the spatial region that has the highest corresponding score compared to other frames. Then, we repeat this operation for every spatial region and concatenate those regions together to obtain a feature map that contains the most discriminative regions of input frames. Next, we use every attention score as a weight and employ the element-wise multiplication on every split feature map to generate another feature map with the global information of input frames. Finally, we concatenate these two feature maps together and employ the global average pooling followed by a fully connected layer to generate the representation vector $X$ for the Re-ID task.
\begin{equation}
X=[x_{1}, x{2}, ..., x_{n}]
\end{equation}

\begin{algorithm}
\footnotesize
\DontPrintSemicolon 
\SetKwInOut{Input}{Input {} {} {} {} {}}
\SetKwInOut{Initialization}{Initialization {} {}}
\Input{A set of feature maps $\{f_n\}_{n=1:N}$;\\ An attention score matrix $S$;\\ Size of Feature map, $H, W, D$; \\ Length of Sequence $N$; \\Number of Spatial blocks $K$;}
\KwOut{Feature map after fusion $F_{fuse}$;}
\Initialization{$F_1 = Zeros(H, W, D)$, $F_2 = Zeros(H, W, D)$;\\ $H_{s} = \lfloor H/4 \rfloor$;}
\For{n = 1 : N}{
	\For{k = 1 : K}{
    	$f_{n,k} \gets f_{n}(H_{s}*(k-1): H_{s}*k, :, :)$;
    }
}
\For{k = 1 : K}{
	$m \gets$ the index of the maximum value in $S(:,k)$; \\
    $F_1(H_{s}: H_{s}, :, :) \gets f_{mk}$; \\
    \For{n = 1:N}{
    	$F_2(H_{s}*(k-1): H_{s}*k, :, :) \mathrel{+}= f_{nk} * s_{nk}$;
    }
}
$F_{fuse} \gets {F_1, F_2}$; \\
\Return{$F_{fuse}$};
\caption{Algorithm of Feature Fusion Strategy}
\label{algo:a1}

\end{algorithm}

\subsection{Loss Function}
In this paper, we utilize both the batch-hard triplet loss proposed in \cite{hermans2017defense} and the softmax loss jointly to train the STA model through the combination the metric learning and discriminative learning.

The triplet loss with hard mining is first proposed in~\cite{hermans2017defense} as an improved version of the original semi-hard triplet loss~\cite{schroff2015facenet}. We randomly sample $P$ identities and $K$ tracklets for each mini-batch to meet the requirement of the batch-hard triplet loss. Typically, the loss function is formulated as follows:
\begin{equation}
\begin{aligned}
L_{triplet} = \sum_{i=1}^P\sum_{a=1}^K[\alpha + &
\overbrace{\max_{p=1...K}||x_{a}^{(i)}-x_{p}^{(i)}||_2}^{hardest \ positive} \\
&- \underbrace{\min_{\substack{n=1...K\\
{j=1...P}\\
j \neq i}} ||x_{a}^{(i)}-x_{p}^{(i)}||_{2}}_{hardest \ negative}]_+
\end{aligned}
\end{equation}
where $x_{a}^{(i)}, x_{p}^{(i)}, x_{n}^{(i)}$ are features extracted from the anchor, positive and negative samples respectively, and $\alpha$ is the margin hyperparameter to control the differences of intra and inter distances. Here, positive and negative samples refer to the person with same or different identity from the anchor.

Besides batch-hard triplet loss, we employ softmax cross entropy loss for discriminative learning as well. The original softmax cross entropy loss can be formulated as follows:
\begin{equation}
    L_{softmax} = -\sum^P_{i=1}\sum^{K}_{a=1} \log \frac{e^{W^T_{y_{a, i}}x_{a,i}}}{\sum_{k=1}^{C}e^{W^T_{k}x_{a,i}}}
\end{equation}
where $y_{i,a}$ is the ground truth identity of the sample $\{a,i\}$, and $C$ is number of classes. Our loss function for optimization is the combination of softmax loss and batch-hard triplet loss as follows:
\begin{equation}\label{eq:loss}
L_{total} = L_{softmax} + L_{triplet}
\end{equation}

\begin{table*}[t]\setlength{\tabcolsep}{8pt}
\centering
\footnotesize
\begin{tabular}{l|c|c|c|c|c|c|c|c}
\hline
\multirow{2}{*}{Model} & \multicolumn{4}{c}{MARS}  & \multicolumn{4}{|c}{DukeMTMC-VideoReID}  \\
\cline{2-9}
& R1 & R5 & R10 & mAP & R1 & R5 & R10 & mAP \\ \hline
Baseline & 74.5 & 88.8 & 91.8 & 64.0 & 79.1 & 93.9 & 96.0 & 76.8 \\
Baseline + TL & 80.8 & 92.1 & 94.3 & 74.0 & 90.6 & 95.8 & 96.7 & 89.7 \\
Baseline + TL + Avg & 82.5 & 92.9 & 94.9 & 75.0 & 91.8 & 97.4 & 98.0 & 91.0 \\
Baseline + TL + STA & 84.8 & 94.6 & 96.2 & 78.0 & 93.3 & 98.1 & 98.6 & 92.7   \\
Baseline + TL + STA + Fusion & 85.3 & 95.1 & 96.4 & 79.1 & 95.3 & 98.1 & 99.1 & 93.9 \\
Baseline + TL + STA + Fusion + Reg & 86.3 & 95.7 & 97.1 & 80.8 & 96.2 & 99.3 & 99.6 & 94.9 \\ \hline
\end{tabular}
\caption{Comparison of different proposed components, where TL, Avg, STA, Fusion, and Reg represent the triplet loss, average pooling, spatial-temporal attention module, feature fusion strategy, and inter-frame regularization respectively. R-1, R-5, R-10 accuracies (\%) and mAP (\%) are reported. Baseline model corresponds to ResNet50 trained with softmax loss on video datasets MARS or DukeMTMC-VideoReID respectively.}
\label{exp:sta}
\vspace{-2mm}
\end{table*}

\begin{table*}[t]\setlength{\tabcolsep}{12pt}
\centering
\footnotesize
\begin{tabular}{c|c|c|c|c|c|c|c|c}
\hline
\multirow{2}{*}{Sequence Length} & \multicolumn{4}{c}{MARS}  & \multicolumn{4}{|c}{DukeMTMC-VideoReID}  \\
\cline{2-9}
 & R1 & R5 & R10 & mAP & R1 & R5 & R10 & mAP \\ \hline
N=2 & 81.7 & 93.8 & 95.7 & 75.7 & 90.3 & 97.6 & 98.6 & 89.0 \\
N=4 & 86.3 & 95.7 & 97.1 & 80.8 & 96.2 & 99.3 & 99.6 & 94.9 \\
N=6 & 86.2 & 95.7 & 96.9 & 81.0 & 96.0 & 99.4 & 99.7 & 95.0 \\
N=8 & 86.2 & 95.7 & 97.1 & 81.2 & 96.0 & 99.3 & 99.6 & 95.0  \\ \hline
\end{tabular}
\caption{Performance comparison of STA model with different sequence lengths during testing on MARS dataset and Duke-VideoReID dataset. Here, we use the model trained with the sequence length of 4 and the spatial regions number of 4.}
\label{exp:seq}
\vspace{-2mm}
\end{table*} 
\section{Experiments}
\subsection{Datasets and Evaluation Protocol}
{\bf Mars dataset}~\cite{zheng2016mars} is one of the largest video-based person re-identification dataset. It contains 17,503 tracklets from 1,261 identities, and additional 3,248 tracklets serving as distractors. These video tracklets are captured by six cameras in a university campus. The total 1,261 identities are split into 625 identities for training and 636 identities for testing. Every identity in the training set has 13 video tracklets on average, and each tracklet has 59 frames on average. The ground truth labels are detected and tracked using the Deformable Part Model (DPM)~\cite{felzenszwalb2008discriminatively} and GMCP tracker~\cite{zamir2012gmcp}.

{\bf DukeMTMC-VideoReID dataset}~\cite{wu2018cvpr_oneshot} is another large-scale benchmark dataset for video-based person Re-ID. It is derived from the DukeMTMC dataset~\cite{ristani2016MTMC}. The DukeMTMC-VideoReID dataset contains 4,832 tracklets from 1,812 identities, and it is split into 702, 702 and 408 identities for training, testing and distraction respectively. In total, it has 369,656 frames of 2,196 tracklets for training, and 445,764 frames of 2,636 tracklets for testing and distraction. Each tracklet has 168 frames on average. The bounding boxes are annotated manually.

{\bf Evaluation Protocol.}
In our experiments, we use the Cumulative Matching Characteristic (CMC) curve and the mean average precision (mAP) to evaluate the performance of the STA model. For each query, CMC represents the accuracy of the person retrieval. We report the Rank-1, Rank-5, Rank-20 scores to represent the CMC curve. The CMC metric is effective when each query corresponds to only one ground truth clip in the gallery. However, when multiple ground truth clips exist in the gallery, and the objective is to return to the user as many correct matches as possible, CMC would not effectively measure the performances of models on this objective. Comparatively, mAP is a comprehensive metric that is well-suited for both single-match and multiple-match objectives.

\subsection{Implementation Details}\label{sec:implement_details}
\begin{table*}[t]\setlength{\tabcolsep}{12pt}
\centering
\footnotesize
\begin{tabular}{c|c|c|c|c|c|c|c|c}
\hline
\multirow{2}{*}{Number of Spatial Regions} & \multicolumn{4}{c}{MARS}  & \multicolumn{4}{|c}{DukeMTMC-VideoReID}  \\
\cline{2-9}
 & R1 & R5 & R10 & mAP & R1 & R5 & R10 & mAP \\ \hline
K=2 & 85.3 & 95.1 & 96.6 & 80.3 & 94.7 & 99.0 & 99.6 & 93.8\\
K=4 & 86.3 & 95.7 & 97.1 & 80.8 & 96.2 & 99.3 & 99.6 & 94.9 \\
K=8 & 85.5 & 95.3 & 96.9 & 80.4 & 95.2 & 99.1 & 99.4 & 93.8 \\ \hline
\end{tabular}
\caption{Performance comparison of the STA model trained with different number of spatial regions on MARS dataset and Duke-VideoReID dataset. Here, we keep the sequence length as a constant of 4.}
\label{exp:spatial}
\end{table*}

\begin{table*}[t]\setlength{\tabcolsep}{12pt}
\centering
\footnotesize
\begin{tabular} {l|c|c|c|c}
\hline
Model & R1 & R5 & R20 & mAP\\ \hline
CNN+Kiss.+MQ~\cite{zheng2016mars} & 68.3 & 82.6 & 89.4 & 49.3 \\
SeeForest~\cite{zhou2017see} & 70.6 & 90.0 & 97.6 & 50.7 \\
Latent Parts~\cite{li2017learning} & 71.8 & 86.6 & 93.0 & 56.1 \\
QAN~\cite{liu2017quality} & 73.7 & 84.9 & 91.6 & 51.7 \\
K-reciprocal~\cite{zhong2017re} & 73.9 & -- & -- & 68.5 \\
TriNet~\cite{hermans2017defense} & 79.8 & 91.4 & -- & 67.7 \\
RQEN~\cite{song2017region} & 77.8 & 88.8 & 94.3 & 71.1 \\
CSACSE~\cite{chen2018video} & 81.2 & 92.1 & -- & 69.4 \\
STAN~\cite{li2018diversity} & 82.3 & -- & -- & 65.8 \\
CSACSE~\cite{chen2018video} + Optical Flow & 86.3 & 94.7 & {\bf 98.2} & 76.1 \\ \hline
STA & {\bf 86.3} & {\bf 95.7} & 98.1 & {\bf 80.8} \\
STA + ReRank & {\bf 87.2} & {\bf 96.2} & {\bf 98.6 } & {\bf 87.7} \\ \hline
\end{tabular}
\caption{Comparison of the STA model with the state-of-the-arts on MARS dataset. Here, we show the results tested with sequence length of 4 and spatial region number of 4 as well.}
\label{exp:mars}
\end{table*}

\begin{table}[t]\setlength{\tabcolsep}{4pt}
\centering
\footnotesize
\begin{tabular} {l|l|l|l|l}
\hline
Model & R1 & R5 & R20 & mAP\\ \hline
ETAP-Net(supervised)~\cite{wu2018cvpr_oneshot} & 83.6 & 94.6 & 97.6 & 78.3 \\ \hline
STA & {\bf 96.2} & {\bf 99.3} & {\bf 99.6} & {\bf 94.9} \\ \hline
\end{tabular}
\caption{Comparison of the STA model with the state-of-the-art on DukeMTMC-VideoReID dataset. Here, we show the results tested with sequence length of 4 and spatial region number of 4 as well.}
\label{exp:duke}
\end{table}

As discussed in ``Proposed Method'', we first randomly select $N=4$ frames from the input tracklet, and use the modified ResNet50 initialized on the ImageNet~\cite{deng2009imagenet} dataset as the backbone network. The number of spatial regions is set to $K=4$. And, each frame is augmented by random horizontal flipping and normalization. Each mini-batch is sampled with randomly selected $P$ identities and randomly sampled $K$ images for each identity from the training set. In our experiment, we set $P=16$ and $K=4$ so that the mini-batch size is $64$. And, we recommend to set the margin parameter in triplet loss to $0.3$. During training, we use the Adam ~\cite{kingma2014adam} weight decay $0.0005$ to optimize the parameters for $70$ epochs. The overall learning rate is initialized to $0.0003$ and decay to $3 \times 10^{-5}$ and $3 \times 10^{-6}$ after training for $200$ and $400$ epochs respectively. The total training process lasts for 800 epochs. For evaluation, we extract the feature vector after the first fully connected layer as the representation of query tracklet.
Our model is implemented on Pytorch platform and trained with two NVIDIA TITAN X GPUs. All our experiments on different datasets follow the same settings as above.

\subsection{Ablation Study}
To verify the effectiveness of each component in STA model, we conduct several analytic experiments including w/ or w/o triplet loss, w/ or w/o spatial-temporal attention model, w/ or w/o feature fusion, and w/ or w/o inter-frame regularization. In addition, we carry out experiments to investigate the effect of varying the sequence length $N$ and the number $K$ of spatial regions. Note that all the remaining settings are the same as those discussed in ``Implementation Details''.

{\bf Effectiveness of Components.}
In Table~\ref{exp:sta}, we list the results of each component in our STA framework. {\bf Baseline} represents the ResNet50 model trained with softmax loss on MARS/DukeMTMC-VideoReID dataset. {\bf TL} corresponds to the hard-batch triplet loss, and ``+ TL'' means the hard-batch triplet loss is combined with the softmax loss in Baseline model. Note that, the {\bf Baseline} and {\bf Baseline + TL} both treat each tracklet frame by frame, i.e., the image-based models. {\bf STA} is our proposed spatial-temporal attention model in which the number of spatial regions is set to $K=4$ and the input sequence length is set to $N=4$ as well. It generates a $4 \times 4$ attention score matrix, and uses the score of the same region but from a different frame to calculate the weighted sum feature maps of each region.
Compared to {\bf Baseline + TL}, {\bf STA} improves Rank-1 and mAP accuracy by $4.0\%$ and $4.0\%$ on MARS, as well as $2.7\%$ and $2.4\%$ on DukeMTMC-VideoReID respectively. These results show that the spatial-temporal attention model is very effective at discovering discriminative image regions which are useful for boosting re-identification performance. {\bf Fusion} means aggregating feature representation by the proposed fusion strategy described in Alg.~\ref{algo:a1}. It is obvious that the proposed fusion strategy can further improve the performance by combining the most discriminative information and global information together. {\bf Reg} refers to the proposed inter-frame regularization term. From the comparison of w/ and w/o {\bf Reg}, we can find the Rank-1 accuracy and mAP improve by $1.0\%$ and $1.7\%$ on Mars, as well as $0.9\%$ and $1.0\%$ on DukeMTMC-VideoReID respectively. This improvement shows the proposed inter-frame regularization term can balance the frame diversity and thus further improve the performance.

{\bf Consistence of Sequence Length.} In Table~\ref{exp:seq}, we show the robustness of the trained model to different sequence length of input tracklet. For fair comparison, we use the model trained with sequence length of 4 and spatial region number of 4, and evaluate the performance with different sequence length: 2, 4, 6, and 8. As we can see, the performances are very consistent when input sequence length is 4, 6 or 8. (\eg all the Rank-1 accuracies are above $85\%$ and the mAP are above $80\%$ on MARS.)
This is because the proposed STA model does not involve more parameters, so there is no restriction on the sequence length. In addition, the performances with $N=4$ surpass most of the state-of-the-art methods which usually need more frames for good representation of the input video.

{\bf Influence of Spatial Region Number.} To investigate how the number of spatial regions influences the final performance, we conduct experiments with three different spatial regions: 2, 4, and 8 with the same sequence length 4. The results are listed in Table~\ref{exp:spatial}. In these experiments, the STA framework always achieves the best result with 4 spatial regions. Since the size of feature maps is $16 \times 8$, with 2 or 8 spatial regions, it could be too coarse to contain enough information or too small to contain enough information for the re-identification task.

\subsection{Comparison with the State-of-the-arts}
Table~\ref{exp:mars} and Table~\ref{exp:duke} report the comparison of our proposed STA model with the state-of-the-art techniques. On each dataset, our approach achieves the best performance, especially on mAP. We attain R1/mAP: $86.3/80.8 (87.2/87.7)$ on MARS before and after re-ranking. In addition, we achieve R1/mAP: $96.2/94.9$ on DukeMTMC-VideoReID.

{\bf Results on MARS.} Comparisons between our approach and the state-of-the-art approaches on MARS are shown in Table~\ref{exp:mars}.
The results show that our approach achieves {\bf 80.8\%} in mAP, which surpasses all existing work by more than {\bf 4.0\%}. Even for the Rank-1 and Rank-5, our approach achieves competitive results compared to the most recent work listed in Table~\ref{exp:mars}. It's noted that the CSACSE + Optical Flow method~\cite{chen2018video} incorporates optical flow as the extra information, which not only brings in more computation, but also causes the drawback that its whole network cannot be trained end-to-end. Comparing to other related work that also does not use optical flow, our approach improves the Rank-1 accuracy and mAP by {\bf 4.0\%} and {\bf 15.0\%} respectively. After implementing re-ranking, the Rank-1 accuracy and mAP can be improved to {\bf 87.2\%} and {\bf 87.7\%}, which outperforms the CSACSE + Optical Flow method~\cite{chen2018video} by {\bf 11.6\%} on mAP.

{\bf Results on DukeMTMC-VideoReID.} Table~\ref{exp:duke} shows the video-base Re-ID performance on DukeMTMC-VideoReID dataset. This dataset is new to the field, and there is only one published baseline~\cite{wu2018cvpr_oneshot} on this dataset. Comparing to this baseline, our approach improves more than {\bf 10\%} on Rank-1 accuracy and mAP. Although there are few results reported on this dataset, we have good reason to believe that our approach works well because it achieves {\bf 96.2\%} and {\bf 94.9\%} on Rank-1 accuracy and mAP respectively.

\section{Conclusion}
This paper addresses the large-scale video-based person re-identification (Re-ID) problem with the proposed Spatial-Temporal Attention (STA) module. In this STA module, instead of directly extracting the video representation through frame-level feature aggregation (\eg average pooling), the 2-D spatial-temporal map is used to calculate the more robust clip-level feature representation without using any additional parameters. The inter-frame regularization and feature fusion strategy are proposed to further improve the clip-level representation. Extensive experiments conducted on two challenging large-scale benchmarks including MARS and DukeMTMC-VideoReID demonstrate the effectiveness of the proposed module for the large-scale video-based person Re-ID problem comparing to the existing methods.

The current approach focuses on the person Re-ID using the provided person tracklets as inputs. A worthy future study would be combining the proposed STA approach with the person detection/tracking algorithms, and apply the combined module in the real-world multi-camera systems.


\vspace{3mm}
\noindent
\textbf{Acknowledgements.}
Yang Fu is supported in part by CloudWalk at UIUC. Yunchao Wei is supported in part by IBM-ILLINOIS Center for Cognitive Computing Systems Research (C3SR) - a research collaboration as part of the IBM AI Horizons Network. Nokia owns the intellectual property rights of this work.
\bibliographystyle{aaai}
\bibliography{aaai}

\begin{thebibliography}{}

\bibitem[\protect\citeauthoryear{Ahmed, Jones, and
  Marks}{2015}]{ahmed2015improved}
Ahmed, E.; Jones, M.; and Marks, T.~K.
\newblock 2015.
\newblock An improved deep learning architecture for person re-identification.
\newblock In {\em {IEEE ICCV}}.

\bibitem[\protect\citeauthoryear{Chen \bgroup et al\mbox.\egroup
  }{2018}]{chen2018video}
Chen, D.; Li, H.; Xiao, T.; Yi, S.; and Wang, X.
\newblock 2018.
\newblock Video person re-identification with competitive snippet-similarity
  aggregation and co-attentive snippet embedding.
\newblock In {\em {IEEE CVPR}}.

\bibitem[\protect\citeauthoryear{Deng \bgroup et al\mbox.\egroup
  }{2009}]{deng2009imagenet}
Deng, J.; Dong, W.; Socher, R.; Li, L.-J.; Li, K.; and Fei-Fei, L.
\newblock 2009.
\newblock Imagenet: A large-scale hierarchical image database.
\newblock In {\em {IEEE CVPR}}.

\bibitem[\protect\citeauthoryear{Ding \bgroup et al\mbox.\egroup
  }{2015}]{ding2015deep}
Ding, S.; Lin, L.; Wang, G.; and Chao, H.
\newblock 2015.
\newblock Deep feature learning with relative distance comparison for person
  re-identification.
\newblock {\em Pattern Recognition}.

\bibitem[\protect\citeauthoryear{Felzenszwalb, McAllester, and
  Ramanan}{2008}]{felzenszwalb2008discriminatively}
Felzenszwalb, P.; McAllester, D.; and Ramanan, D.
\newblock 2008.
\newblock A discriminatively trained, multiscale, deformable part model.
\newblock In {\em {IEEE CVPR}}.

\bibitem[\protect\citeauthoryear{Fu \bgroup et al\mbox.\egroup
  }{2019}]{fu2018horizontal}
Fu, Y.; Wei, Y.; Zhou, Y.; Shi, H.; Huang, G.; Wang, X.; Yao, Z.; and Huang, T.
\newblock 2019.
\newblock Horizontal pyramid matching for person re-identification.
\newblock In {\em AAAI}.

\bibitem[\protect\citeauthoryear{He \bgroup et al\mbox.\egroup
  }{2016}]{he2016deep}
He, K.; Zhang, X.; Ren, S.; and Sun, J.
\newblock 2016.
\newblock Deep residual learning for image recognition.
\newblock In {\em {IEEE CVPR}}.

\bibitem[\protect\citeauthoryear{Hermans, Beyer, and
  Leibe}{2017}]{hermans2017defense}
Hermans, A.; Beyer, L.; and Leibe, B.
\newblock 2017.
\newblock In defense of the triplet loss for person re-identification.
\newblock {\em arXiv preprint arXiv:1703.07737}.

\bibitem[\protect\citeauthoryear{Hirzer \bgroup et al\mbox.\egroup
  }{2011}]{hirzer2011person}
Hirzer, M.; Beleznai, C.; Roth, P.~M.; and Bischof, H.
\newblock 2011.
\newblock Person re-identification by descriptive and discriminative
  classification.
\newblock In {\em SCIA}.

\bibitem[\protect\citeauthoryear{Jimin~Xiao}{2019}]{xiao2019ian}
Jimin~Xiao, Yanchun~Xie, T. T. K. H. Y. W. J.~F.
\newblock 2019.
\newblock Ian: The individual aggregation network for person search.
\newblock {\em Pattern Recognition}.

\bibitem[\protect\citeauthoryear{Kingma and Ba}{2014}]{kingma2014adam}
Kingma, D.~P., and Ba, J.
\newblock 2014.
\newblock Adam: A method for stochastic optimization.
\newblock {\em arXiv preprint arXiv:1412.6980}.

\bibitem[\protect\citeauthoryear{Li \bgroup et al\mbox.\egroup
  }{2017}]{li2017learning}
Li, D.; Chen, X.; Zhang, Z.; and Huang, K.
\newblock 2017.
\newblock Learning deep context-aware features over body and latent parts for
  person re-identification.
\newblock In {\em {IEEE CVPR}}.

\bibitem[\protect\citeauthoryear{Li \bgroup et al\mbox.\egroup
  }{2018}]{li2018diversity}
Li, S.; Bak, S.; Carr, P.; and Wang, X.
\newblock 2018.
\newblock Diversity regularized spatiotemporal attention for video-based person
  re-identification.
\newblock In {\em {IEEE CVPR}}.

\bibitem[\protect\citeauthoryear{Lin \bgroup et al\mbox.\egroup
  }{2017}]{lin2017structured}
Lin, Z.; Feng, M.; Santos, C. N.~d.; Yu, M.; Xiang, B.; Zhou, B.; and Bengio,
  Y.
\newblock 2017.
\newblock A structured self-attentive sentence embedding.
\newblock {\em arXiv preprint arXiv:1703.03130}.

\bibitem[\protect\citeauthoryear{Liu \bgroup et al\mbox.\egroup
  }{2017}]{liu2017video}
Liu, H.; Jie, Z.; Jayashree, K.; Qi, M.; Jiang, J.; Yan, S.; and Feng, J.
\newblock 2017.
\newblock Video-based person re-identification with accumulative motion
  context.
\newblock {\em IEEE TCSVT}.

\bibitem[\protect\citeauthoryear{Liu, Yan, and Ouyang}{2017}]{liu2017quality}
Liu, Y.; Yan, J.; and Ouyang, W.
\newblock 2017.
\newblock Quality aware network for set to set recognition.
\newblock In {\em {IEEE CVPR}}.

\bibitem[\protect\citeauthoryear{{McLaughlin}, {Martinez del Rincon}, and
  {Miller}}{2016}]{mclaughlin2016recurrent}
{McLaughlin}, N.; {Martinez del Rincon}, J.; and {Miller}, P.
\newblock 2016.
\newblock Recurrent convolutional network for video-based person
  re-identification.
\newblock In {\em {IEEE CVPR}}.

\bibitem[\protect\citeauthoryear{Meyer}{2000}]{meyer2000matrix}
Meyer, C.~D.
\newblock 2000.
\newblock {\em Matrix analysis and applied linear algebra}.
\newblock Siam.

\bibitem[\protect\citeauthoryear{Ristani \bgroup et al\mbox.\egroup
  }{2016}]{ristani2016MTMC}
Ristani, E.; Solera, F.; Zou, R.; Cucchiara, R.; and Tomasi, C.
\newblock 2016.
\newblock Performance measures and a data set for multi-target, multi-camera
  tracking.
\newblock In {\em {ECCV}}.

\bibitem[\protect\citeauthoryear{Schroff, Kalenichenko, and
  Philbin}{2015}]{schroff2015facenet}
Schroff, F.; Kalenichenko, D.; and Philbin, J.
\newblock 2015.
\newblock Facenet: A unified embedding for face recognition and clustering.
\newblock In {\em {IEEE CVPR}}.

\bibitem[\protect\citeauthoryear{Simonyan and
  Zisserman}{2014}]{simonyan2014very}
Simonyan, K., and Zisserman, A.
\newblock 2014.
\newblock Very deep convolutional networks for large-scale image recognition.
\newblock {\em arXiv preprint arXiv:1409.1556}.

\bibitem[\protect\citeauthoryear{Song \bgroup et al\mbox.\egroup
  }{2017}]{song2017region}
Song, G.; Leng, B.; Liu, Y.; Hetang, C.; and Cai, S.
\newblock 2017.
\newblock Region-based quality estimation network for large-scale person
  re-identification.
\newblock {\em arXiv preprint arXiv:1711.08766}.

\bibitem[\protect\citeauthoryear{Sun \bgroup et al\mbox.\egroup
  }{2017}]{sun2017beyond}
Sun, Y.; Zheng, L.; Yang, Y.; Tian, Q.; and Wang, S.
\newblock 2017.
\newblock Beyond part models: Person retrieval with refined part pooling.
\newblock {\em arXiv preprint arXiv:1711.09349}.

\bibitem[\protect\citeauthoryear{Szegedy \bgroup et al\mbox.\egroup
  }{2016}]{szegedy2016rethinking}
Szegedy, C.; Vanhoucke, V.; Ioffe, S.; Shlens, J.; and Wojna, Z.
\newblock 2016.
\newblock Rethinking the inception architecture for computer vision.
\newblock In {\em {IEEE CVPR}}.

\bibitem[\protect\citeauthoryear{Wang \bgroup et al\mbox.\egroup
  }{2014}]{wang2014person}
Wang, T.; Gong, S.; Zhu, X.; and Wang, S.
\newblock 2014.
\newblock Person re-identification by video ranking.
\newblock In {\em {ECCV}}.

\bibitem[\protect\citeauthoryear{Wang \bgroup et al\mbox.\egroup
  }{2016}]{wang2016person}
Wang, T.; Gong, S.; Zhu, X.; and Wang, S.
\newblock 2016.
\newblock Person re-identification by discriminative selection in video
  ranking.
\newblock {\em {IEEE TPAMI}}.

\bibitem[\protect\citeauthoryear{Wu \bgroup et al\mbox.\egroup
  }{2018}]{wu2018cvpr_oneshot}
Wu, Y.; Lin, Y.; Dong, X.; Yan, Y.; Ouyang, W.; and Yang, Y.
\newblock 2018.
\newblock Exploit the unknown gradually: One-shot video-based person
  re-identification by stepwise learning.
\newblock In {\em {IEEE CVPR}}.

\bibitem[\protect\citeauthoryear{Xiao \bgroup et al\mbox.\egroup
  }{2016}]{xiao2016learning}
Xiao, T.; Li, H.; Ouyang, W.; and Wang, X.
\newblock 2016.
\newblock Learning deep feature representations with domain guided dropout for
  person re-identification.
\newblock In {\em {IEEE CVPR}}.

\bibitem[\protect\citeauthoryear{Xu \bgroup et al\mbox.\egroup
  }{2015}]{xu2015show}
Xu, K.; Ba, J.; Kiros, R.; Cho, K.; Courville, A.; Salakhudinov, R.; Zemel, R.;
  and Bengio, Y.
\newblock 2015.
\newblock Show, attend and tell: Neural image caption generation with visual
  attention.
\newblock In {\em {ICML}}.

\bibitem[\protect\citeauthoryear{Xu \bgroup et al\mbox.\egroup
  }{2017}]{xu2017jointly}
Xu, S.; Cheng, Y.; Gu, K.; Yang, Y.; Chang, S.; and Zhou, P.
\newblock 2017.
\newblock Jointly attentive spatial-temporal pooling networks for video-based
  person re-identification.
\newblock {\em arXiv preprint arXiv:1708.02286}.

\bibitem[\protect\citeauthoryear{You \bgroup et al\mbox.\egroup
  }{2016}]{you2016top}
You, J.; Wu, A.; Li, X.; and Zheng, W.-S.
\newblock 2016.
\newblock Top-push video-based person re-identification.
\newblock In {\em {IEEE CVPR}}.

\bibitem[\protect\citeauthoryear{Zamir, Dehghan, and
  Shah}{2012}]{zamir2012gmcp}
Zamir, A.~R.; Dehghan, A.; and Shah, M.
\newblock 2012.
\newblock Gmcp-tracker: Global multi-object tracking using generalized minimum
  clique graphs.
\newblock In {\em {ECCV}}.

\bibitem[\protect\citeauthoryear{Zheng \bgroup et al\mbox.\egroup
  }{2016}]{zheng2016mars}
Zheng, L.; Bie, Z.; Sun, Y.; Wang, J.; Su, C.; Wang, S.; and Tian, Q.
\newblock 2016.
\newblock Mars: A video benchmark for large-scale person re-identification.
\newblock In {\em {ECCV}}.

\bibitem[\protect\citeauthoryear{Zheng, Zheng, and
  Yang}{2017}]{zheng2017discriminatively}
Zheng, Z.; Zheng, L.; and Yang, Y.
\newblock 2017.
\newblock A discriminatively learned cnn embedding for person reidentification.
\newblock {\em ACM TOMM}.

\bibitem[\protect\citeauthoryear{Zhong \bgroup et al\mbox.\egroup
  }{2017}]{zhong2017re}
Zhong, Z.; Zheng, L.; Cao, D.; and Li, S.
\newblock 2017.
\newblock Re-ranking person re-identification with k-reciprocal encoding.
\newblock In {\em {IEEE CVPR}}.

\bibitem[\protect\citeauthoryear{Zhou \bgroup et al\mbox.\egroup
  }{2017}]{zhou2017see}
Zhou, Z.; Huang, Y.; Wang, W.; Wang, L.; and Tan, T.
\newblock 2017.
\newblock See the forest for the trees: Joint spatial and temporal recurrent
  neural networks for video-based person re-identification.
\newblock In {\em {IEEE CVPR}}.

\end{thebibliography}
\end{document}